\documentclass[journal]{IEEEtran}

\usepackage{times}
\usepackage{epsfig}
\usepackage{graphicx}
\usepackage{amsmath}
\usepackage{amssymb}
\usepackage{pdfpages}
\usepackage{algorithm, algpseudocode}
\usepackage{color}
\definecolor{citecolor}{RGB}{119,185,0} 
\usepackage[pagebackref=false,breaklinks=true,letterpaper=true,colorlinks,citecolor=citecolor,bookmarks=false]{hyperref}
\usepackage{colortbl}

\usepackage{pifont}

\usepackage{amsmath}

\DeclareMathOperator*{\E}{\mathbb{E}}
\def\eg{\emph{e.g.}} 
\def\ie{\emph{i.e.}} 
\def\etal{\emph{et~al.}} 

\newlength\savewidth\newcommand\shline{\noalign{\global\savewidth\arrayrulewidth
  \global\arrayrulewidth 1pt}\hline\noalign{\global\arrayrulewidth\savewidth}}

\begin{document}

%%%%%%%%% TITLE
\title{Adaptive Boosting for Domain Adaptation: 
\\Towards Robust Predictions in Scene Segmentation}

\author{Zhedong Zheng and Yi Yang,~\IEEEmembership{Senior~Member,~IEEE} 
\thanks{Zhedong Zheng is with the School of Computing, National University of Singapore, Singapore 118404. E-mail: zdzheng@nus.edu.sg}
\thanks{
Yi Yang is with the College of Computer Science and Technology, Zhejiang
University, China 310027. E-mail: yangyics@zju.edu.cn. He is in part supported by the Fundamental Research Funds for the Central Universities (No. 226-2022-00087). }

\thanks{Yi Yang is the corresponding author.}
}

% The paper headers
\markboth{Journal of \LaTeX\ Class Files,~Vol.~14, No.~8, August~2015}%
{Shell \MakeLowercase{\textit{et al.}}: Bare Demo of IEEEtran.cls for IEEE Journals}

\maketitle

%%%%%%%%% ABSTRACT
\begin{abstract}
Domain adaptation is to transfer the shared knowledge learned from the source domain to a new environment, \ie, target domain. One common practice is to train the model on both labeled source-domain data and unlabeled target-domain data. Yet the learned models are usually biased due to the strong supervision of the source domain. Most researchers adopt the early-stopping strategy to prevent over-fitting, but when to stop training remains a challenging problem since the lack of the target-domain validation set. 
In this paper, we propose one efficient bootstrapping method, called Adaboost Student, explicitly learning complementary models during training and liberating users from empirical early stopping. Adaboost Student combines deep model learning with the conventional training strategy, \ie, adaptive boosting, and enables interactions between learned models and the data sampler.
We adopt one adaptive data sampler to progressively facilitate learning on hard samples and aggregate ``weak'' models to prevent over-fitting. 
Extensive experiments show that (1) Without the need to worry about the stopping time, AdaBoost Student provides one robust solution by efficient complementary model learning during training. 
(2) AdaBoost Student is orthogonal to most domain adaptation methods, which can be combined with existing approaches to further improve the state-of-the-art performance. We have achieved competitive results on three widely-used scene segmentation domain adaptation benchmarks.  
\end{abstract}


%%%%%%%%% BODY TEXT
\section{Introduction}
\IEEEPARstart{I}{n} recent years, deep learning approaches have achieved significant improvement in many computer vision fields, including  semantic segmentation~\cite{zhao2017pyramid,chen2017deeplab}. However, improving the scalability of deeply-learned models remains a challenging task. For instance, the segmentation models learned on the data collected on sunny days usually perform terribly in different environments, such as rainy days and foggy days~\cite{wu2019ace, wang2019learning}. One straightforward idea is to collect more training data for the target environment and re-train one domain-specific model for inference. However, it is usually unaffordable to annotate large-scale datasets for every target scenario, especially for the tasks demanding pixel-wise annotations, \eg, scene segmentation.  Therefore, researchers resort to domain adaptation techniques~\cite{wang2019learning,pan2019transferrable,xiang2020unsupervised} to ``borrow'' the common knowledge from homogeneous datasets, \eg, labeled source-domain data. The source-domain data can be the synthetic data generated by game engines, such as GTA5~\cite{xiang2020unsupervised, richter2016playing}, or the large-scale data collected in other real-world scenarios~\cite{cordts2016cityscapes}. In this way, we can largely save the annotation cost as well as the training time to achieve competitive results in the target domain. The common practice is to train the segmentation model on both labeled source-domain data and unlabeled target-domain data. Most existing works focus on domain alignment, minimizing the gap between the source domain and the target domain~\cite{chen2019learning,chen2019crdoco,vu2019dada}. Several early attempts leverage generative models to transfer the image style at the pixel level~\cite{hoffman2018cycada,wu2018dcan}, while other works concentrate on narrowing the semantic gap in the feature space~\cite{geng2011daml}. For instance, Tsai~\etal~\cite{tsai2018learning,tsai2019domain} and Luo~\etal~\cite{luo2019significance,luo2019taking} harness the adversarial loss to encourage the segmentation model to learn domain-invariant features.  

\begin{figure}[t]
\begin{center}
\includegraphics[width=1.0\linewidth]{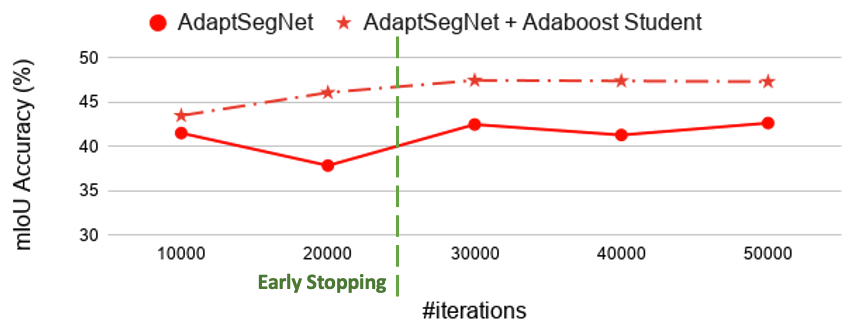}
\end{center}
\vspace{-.2in}
   \caption{The sensitivity analysis of the model on the GTA5~\cite{richter2016playing} $\rightarrow$ Cityscapes~\cite{cordts2016cityscapes} benchmark in the first $50000$ iterations on the target-domain test set. We re-implement the widely-used method AdaptSegNet~\cite{tsai2018learning}, and observe the prediction fluctuation at different training iterations. The proposed Adaboost Student makes the model free from the early-stopping trick and can efficiently converge to one competitive performance on the target domain. For instance, the empirical early-stopping time is usually set at the 25,000-th iteration~\cite{zheng2019unsupervised}, which is  sub-optimal. }
\label{fig:motivation}
\end{figure}

However, one inherent problem exists in the scene segmentation domain adaptation. Due to the strong supervision on the source-domain label, we observe that the model is still prone to over-fitting the source domain, leading to the unstable prediction on the target domain (see Figure~\ref{fig:motivation}). Although several researchers adopt the early-stopping strategy to prevent over-fitting~\cite{luo2019significance,luo2019taking,zheng2019unsupervised}, when to stop the training remains challenging. 
%Furthermore, due to limited GPU memory, extremely small batch size leads to different training orders within every epoch, affecting the consistency of the model snapshots in different epochs. 
The unstable prediction not only compromises the method re-implementation, but also harms users to select the final model in the real-world application. 

In an attempt to overcome the above-mentioned challenge, this paper adopts the spirit of one conventional training strategy, \ie, adaptive boosting~\cite{freund1996experiments}, with the deeply-learned model to facilitate learning complementary models and progressively improve the model scalability. 
In particular, we gradually provide more sampling probability to the ``hard'' samples that achieve high prediction variance in the current model. In this way, we can efficiently obtain several complementary models during one-time training rather than training multiple independent models as the conventional adaptive boosting~\cite{freund1996experiments}. 
Aggregating such model snapshots provide more robust predictions on unlabeled target domain data and, more importantly, prevents the model from over-fitting. 
%As one minor contribution, we also introduce a new regularization term, called, long-term prediction consistency for unlabeled data, which stabilizes the model prediction. 
As a result, we observe a consistent improvement over other domain adaptation approaches on three prevailing benchmarks. 
To summarize, our contributions are: 
\begin{itemize}
    \item %We observe unstable predictions in scene segmentation domain adaptation. 
    an efficient bootstrapping method for scene segmentation domain adaptation, AdaBoost Student, which enables interactions between learned models and the data sampler, and aggregates the ``weak'' model snapshots to prevent over-fitting. AdaBoost Student makes users free from the empirical early stopping trick; and
    \item an adaptive data sampler to progressively increase the sampling probability of hard samples with ambiguous predictions, facilitating the complementary model learning. The criterion of hard sample selection is based on prediction variance in unlabeled target-domain data; and 
    \item a demonstration that the proposed approach has a consistent improvement over existing methods on two synthetic-to-real benchmarks and one cross-city benchmark.
\end{itemize}
The rest of this paper is organized as follows. Section~\ref{sec:relatedwork} discusses relevant works, including semantic segmentation adaptation, bootstrapping learning, and hard example mining. Section~\ref{sec:method} elaborates on the proposed AdaBoost Student approach followed by experiment results in Section~\ref{sec:experiment}. We add further ablation studies and discussion in Section~\ref{sec:further}. Finally, Section~\ref{sec:conclusion} concludes the paper.

\section{Related Work}~\label{sec:relatedwork}
\subsection{Semantic Segmentation Adaptation}
There are two big families of segmentation adaptation. One line of works~\cite{tsai2018learning,tsai2019domain,luo2019significance,luo2020adversarial,gong2019dlow,du2019ssf,chen2019crdoco} concentrates on the domain alignment.
Some pioneering works~\cite{hoffman2018cycada,wu2018dcan} leverage the CycleGAN~\cite{CycleGAN2017} to transfer the image style to the target domain, minimizing the biases from the source domain.
Taking one step further, Wu~\etal~\cite{wu2019ace} and Yue~\etal~\cite{yue2019domain} transfer the input images into different styles, such as rainy and foggy, to learn domain-invariant features. In contrast, several works do not change the image style at the pixel level but leverage the adversarial loss to disentangle the shared knowledge and minimize the negative impact of the domain-specific information. For instance, Tsai~\etal~\cite{tsai2018learning} and Luo~\etal~\cite{luo2019taking} demand the generator to fool the discriminator by discriminating the feature between the source domain and target domain. Sankaranarayanan~\etal~\cite{sankaranarayanan2018learning} leverage the image reconstruction as the self-supervision task to align the two domains, \ie, the real-world environment and synthetic data, yielding a scalable learned model, but it also costs more computation resources for recovering the large-size input image.  
Another line of works~\cite{zou2018unsupervised,zou2019confidence,zheng2019unsupervised,zhou2020uncertainty,li2019bidirectional,zhang2021prototypical} focuses on mining the domain-specific knowledge in the target domain. The main idea is to minimize the prediction entropy in the target domain~\cite{huang2018multi,ding2019feature,DingFXY20}. Zou~\etal~\cite{zou2018unsupervised,zou2019confidence}
propose to leverage the pseudo labels with high confidence, and discuss different regularization terms.
Taking one step further, Feng~\etal~\cite{feng2021complementary} and Lin~\etal~\cite{lin2020unsupervised} leverage the group information to acquire pseudo labels with high precision. Zheng~\etal~\cite{zheng2019unsupervised} apply one two-step training strategy, which first conducts the domain alignment to generate better pseudo labels according to multi-level features~\cite{yang2021multiple}, and then applies the pseudo label learning. Furthermore, Zheng~\etal~\cite{zheng2020unsupervised}  rectify the pseudo label learning by explicitly combining the prediction uncertainty into the cross-entropy loss.
In this paper, we notice that existing works, to some extent, still suffer from over-fitting and early-stopping tricks, leading to large prediction variance and re-implementation difficulties. We propose an efficient bootstrapping method, which is orthogonal to  most existing approaches on minimizing the domain gap. In fact, the proposed method can be fused with existing methods to further improve the model scalability.   

\subsection{Bootstrapping Learning}
Bootstrapping learning denotes the capability of the model to progressively improve the performance by teaching itself~\cite{reed2014training}. The teacher-student training policy~\cite{hinton2015distilling} is first proposed for model distillation, which intends to distill the knowledge from large-scale models to mobile models. It demands the parameter-efficient student model to predict consistent prediction scores with the large teacher model. 
However, teacher-student policy needs one pre-trained sophisticated network as the teacher model in advance, which is not always available. To address this limitation, Zhang~\etal~\cite{zhang2018deep} introduce deep mutual learning to simultaneously train two models of different structures from scratch and learn from each other. 
Different from preserving two models during training, Laine  \etal~\cite{laine2016temporal} propose $\Pi$ model to enable the model learning from the model itself, which demands consistent prediction on the original input and the augmented input. Furthermore, Laine~\etal~also explore the temporal ensembling, which leverages the prediction history to stabilize the training.
To address the out-of-the-date prediction in history, Tarvainen~\etal~\cite{tarvainen2017mean} and Choi~\etal~\cite{choi2019self} propose to maintain one mean model by moving average weight
~\cite{cha2021domain}, which does not need to train one new model again. Similarly, Chen~\etal~\cite{chen2018semi} and Zheng~\etal~\cite{zheng2019unsupervised} introduce the memory mechanism to help the feature update, facilitating consistency learning. 
We are mainly different from existing bootstrapping learning approaches in two aspects: 1) We do not introduce more prerequisites, \eg, extra memory modules or an independent teacher model. Instead, we gradually aggregate the complementary model during training. The model can converge efficiently, and achieve competitive performance in the target domain quickly; 
2) We explicitly consider the model weakness and adopt one adaptive data sampler, which focuses on the ``hard'' samples and facilitates learning complementary models in a coherent manner.

\subsection{Hard Example Mining} 
The neural networks usually underestimate the minority, \ie, hard examples, during the mini-batch optimization~\cite{lin2017focal}, leading to over-fitting most ``easy'' samples. Therefore, several researchers resort to  example mining strategies to add more emphasis on the example with large losses. Most existing works focus on the object detection task. For instance, Shrivastava~\etal~\cite{shrivastava2016training} let the network forward twice, called OHEM, to mine hard samples. OHEM searches hard examples from a large pool of 4000 regions in the first round, and then optimize the network by the selected samples, which are the top 128 high-loss regions. To evade the two-stage forward process,  Lin~\etal~\cite{lin2017focal} propose to directly enlarge the punishment on the hard example by a modified cross entropy loss, which also yields performance improvement. 
In contrast, many metric learning methods~\cite{oh2016deep,hermans2017defense} do not directly modify the loss, but select the input triplets with the largest loss, such as the farthest positives and the closet negatives to help the model learn the distance (relation) in the semantic space. In this work, we do not have the target-domain label, and could not foreknow the wrong samples. Therefore, we resort to the prediction variance as the hardness indicator. In fact, the proposed method is orthogonal to existing work, including focal loss and OHEM (using cross entropy as the hardness indicator). There are three main differences between our work and OHEM: (1) OHEM needs to feed forward the network twice in every iteration, while the proposed method conducts the inference once. We only update the data sampler after every epoch, which is relatively efficient. (2) OHEM is based on cross-entropy loss with ground-truth labels. In our work, we do not have the target-domain ground-truth labels. Therefore, we resort to the prediction variance, and as shown in Table~\ref{table:ablation}, the prediction variance surpasses the prediction entropy.  (3) We further leverage the diversity between different epochs to conduct student aggregation. In the experiment, we also add ablation studies on the compatibility with existing methods.  

\begin{figure}[t]
\begin{center}
\includegraphics[width=1.0\linewidth]{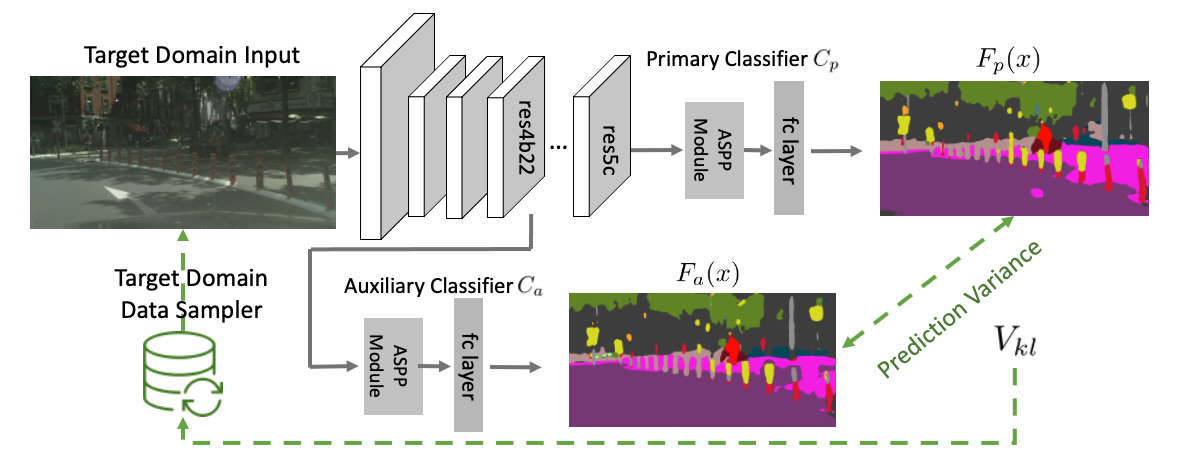}
\end{center}
\vspace{-.2in}
\caption{Illustration of the basic model structure (Gray Parts). We follow most existing works~\cite{tsai2018learning,luo2019significance,luo2019taking,tsai2019domain,zheng2019unsupervised,zheng2020unsupervised} to deploy the modified Deeplab-v2~\cite{chen2017deeplab} with ResNet-101~\cite{he2016deep} as backbone. The auxiliary classifier is added to help the optimization, preventing gradient vanishment, especially for the lower layers~\cite{szegedy2015going, zhao2017pyramid}. In this work, we focus on the outer loop (\textcolor{citecolor}{Green} Parts). We leverage the discrepancy of the segmentation outputs from the primary classifier $C_p$ and $C_a$ to acquire the prediction variance $V_{kl}$. The variance is then utilized to update the data sampler. More details can be found in Section~\ref{sec:sampler}.}
\label{fig:structure}
\end{figure}

\section{Method}~\label{sec:method}
\noindent\textbf{Formulation.} Given the labeled source-domain dataset $X_m = \{x^i_m\}^M_{i=1}$ and the unlabeled target-domain  dataset $X_n = \{x^j_n\}^N_{j=1}$, scene segmentation domain adaptation is to learn the projection function $F$, which maps the input data $X$ to the segmentation map $Y$. $M$ and $N$ denote the number of the source-domain images and the target-domain images, respectively. 
Following the common practice in~\cite{tsai2018learning,luo2019significance,zheng2019unsupervised}, we adopt the modified DeepLabv2~\cite{chen2017deeplab} as the baseline model, which contains two classifiers, \ie, the primary classifier $C_p$ and the auxiliary classifier $C_a$. The auxiliary classifier design is to help optimization and prevent the gradient from vanishing~\cite{szegedy2015going, zhao2017pyramid}. In this work, we also involve the auxiliary prediction into the calculation of the prediction variance $V_{kl}$. To simplify, we denote the two segmentation functions $F_p$ and $F_a$, where $F_p(x)$ is the output of the primary classifier, and $F_a(x)$ is the output of the auxiliary classifier (see Figure~\ref{fig:structure}).

\begin{figure}[t]
\begin{center}
\includegraphics[width=1.0\linewidth]{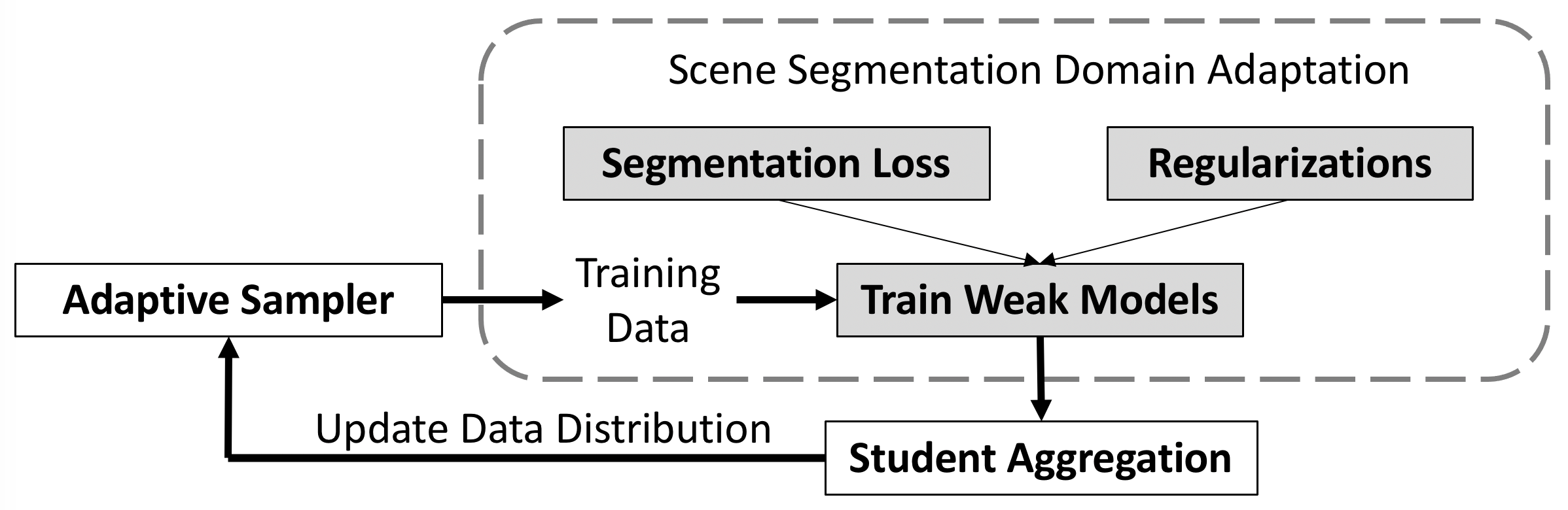}
\end{center}
\vspace{-.2in}
   \caption{ The brief pipeline of the proposed method. There are two main components, \ie, Adaptive Sampler and Student Aggregation. We modify the training data distribution to learn complementary ``weak'' models, preventing the model from over-fitting. \textbf{Different from existing methods, the pipeline enables interactions between learned models and the data sampler. } The proposed method is orthogonal to most existing scene segmentation domain adaptation approaches (in the rounded rectangle). }
\label{fig:pipeline}
\end{figure}

\noindent\textbf{Overview.} In this work, we do not pursue a sophisticated segmentation structure or optimization losses. We focus on the learning strategy. As shown in Figure~\ref{fig:pipeline}, we provide one brief pipeline to illustrate the proposed approach, AdaBoost Student.  AdaBoost Student contains two primary components, \ie, Adaptive Sampler and Student Aggregation. Adaptive Sampler modifies the input training data distribution, and provides more emphasis on uncertain samples. On the other hand, Student Aggregation is to accumulate the learned ``weak'' student models and output the prediction uncertainty to update data distribution. It is worth noting that there is one interaction loop from Adaptive Sampler to Student Aggregation, and back to Adaptive Sampler. 
We adopt one training policy similar to Generative Adversarial Network (GAN)~\cite{goodfellow2014generative}.
During training, Student Aggregation updates model parameters when the input data distribution is fixed, and vice versa. We fix the aggregated student model parameter to update the data distribution of Adaptive Sampler.  We will illustrate the two components extensively in the following sections. 

\subsection{Adaptive Sampler} \label{sec:sampler}
Most bootstrapping works~\cite{reed2014training,laine2016temporal} focus on mining the knowledge from the fixed data distribution but ignore the attention to the ``hard'' samples. The straightforward solution underpinning the conventional method, \ie, Adaptive boosting~\cite{freund1996experiments}, is to improve the sampling rate of the wrong-classified samples. 
In a similar spirit, we adopt one adaptive sampler to progressively update the input data distribution $D_t$ according to the epoch $t$. The initial data distribution $D_1(j) = 1/N$, and every sample has equal sampling probability. To learn the complementary models for the target domain, we improve the sampling rate of the ``hard'' target-domain samples with ambiguous predictions during training. It is worth noting that, due to the missing label in the target-domain, 
we can not deploy the prediction error to find the ``hard'' samples. Instead, we deploy the prediction variance as the ``hard'' indicator. Specifically, we formulate the prediction variance as the KL-divergence between the prediction of the primary classifier and the auxiliary classifier. The prediction variance of one single target-domain image can be calculated as:
\begin{equation}
V_{kl}(x_n^j|\theta) = \E [F_{p}(x_n^j|\theta) \log (\frac{F_{p}(x_n^j|\theta)}{F_{a}(x_n^j|\theta)})].
\label{eq:variance}
\end{equation}
If two classifiers predict the same class prediction, the value $V_{kl}$ is low. Otherwise, the prediction variance is high, which indicates the model is not confident on such training data. In practice, since the output prediction is the pixel-wise segmentation map, we adopt the mean value of the whole prediction map as the indicator for the input image. 
To make $V_{kl}$ as one distribution for the whole dataset, we apply the softmax function to normalize the prediction variance and make the sum of the $V_{kl}$ equal to 1. The next data sampling distribution is averaged with the last data distribution and the normalized prediction variance as: 
\begin{equation}
D_{t+1}(j) = \frac{1}{2}(D_{t}(j) + V_{kl}(x_n^j|\theta)).
\label{eq:distribution}
\end{equation}
%where $Z_t$ is the normalization factor, 
%\begin{equation}
%Z_t = \sum_{j=1}^N (V_{kl}(x^j_n)).
%\end{equation}
Since $\sum_{j=1}^N D_t(j) = 1$ and $\sum_{j=1}^N V_{kl} =1$, we ensure the update distribution $\sum_{j=1}^N D_{t+1} = 1$, and $D_{t+1}(j)$ is positive.
As a result, the samples obtaining high prediction variance have more chances to be sampled in the next epoch. 

\noindent\textbf{Discussion.} 
\textbf{1. Why is there a discrepancy between the primary classifier and the auxiliary classifier?} It is mostly due to the different receptive fields of the two classifiers. Specifically, the input feature of the primary classifier is ``res5c'', while the auxiliary classifier learns from the ``res4b22'' of the ResNet-101 backbone~\cite{he2016deep}. We leverage the prediction variance to represent the model uncertainty on the input data as~\cite{zheng2020unsupervised}. Besides, we insert the dropout function before classifiers, which also enlarges the prediction discrepancy and helps to obtain the accurate uncertainty~\cite{gal2016dropout}. 
\textbf{2. What are the advantages of the proposed adaptive sampler?} There are two main points. First, we do not need the target-domain label to estimate the prediction error, which is different from the conventional AdaBoost algorithm in supervised learning. Instead, we leverage the prediction uncertainty to find the ``hard'' samples, which are more feasible for the domain adaptation problem. 
Second, existing methods usually ignore the importance of the input data distribution. The ``easy'' target-domain data, which  has high prediction confidence, generally does not provide more domain-specific information about the target domain. The adaptive sampler explicitly introduces the prediction variance into the data distribution update (see Eq.~\ref{eq:distribution}). We put more and more emphasis on the data with high uncertainty, facilitating the complementary model learning. The dynamic input distribution via the adaptive sampler encourages the knowledge transfer to the target domain efficiently.  
%It is worth noting that we only modify the data distribution of the target-domain data $X_n$.   
\textbf{3. Other sampling criterion.} Except for the prediction variance, prediction entropy is another alternative choice. In the ablation study, we also provide the empirical result based on the sampling criterion of the prediction entropy, which can be formulated as: $V_{en}(x_n^j|\theta) = \E [-F_{p}(x_n^j|\theta) \log (F_{p}(x_n^j|\theta))]$. A high value indicates that the prediction is close to the uniform distribution, and the model is not confident with the predicted classes given the input data. %In practice, we replace the $V_{kl}$ in Eq.~\ref{eq:distribution} with $V_{en}$ to update the sampler. Similarly,
The samples obtaining high prediction entropy have more chances to be sampled in the next epoch. However, we observe one similar phenomenon with~\cite{zheng2020unsupervised} that the prediction entropy is sensitive to the object edges in segmentation maps rather than the ``hard'' samples with ambiguous predictions, which need to be ``seen'' again. Therefore, the sampling criterion based on prediction entropy generally performs worse than the prediction variance in Eq.~\ref{eq:variance}, which is verified in Table~\ref{table:ablation}.

\subsection{Student Aggregation} \label{sec:aggregation}
In the conventional strategy, \ie, Adaptive Boosting, multiple binary classifiers are sequentially learned with more attentions to the hard samples. Instead of training multiple deeply-learned models sequentially, we leverage the adaptive data sampler to enable learning complementary models in one-time training, which largely saves the training cost. Next, we combine the vote of ``weak'' model snapshots as one final model. %The traditional AdaBoost mainly studies the binary classification problem.
We regard snapshots as student models but do not introduce extra teachers. For the deeply-learned models, it is not efficient to preserve all the model snapshots in the memory, and fuse the prediction of every model, which demands the multiple-time inference. We follow Mean Teacher~\cite{tarvainen2017mean} to apply the weight average moving to keep one weight-averaged model of previous epochs:
\begin{equation}
    \Theta_t = \sum \alpha_t \times \theta_t,
\end{equation}
where the model weight $\alpha_t$ assigns different emphasis to  ``weak'' models, and $\theta_t$ denotes the parameters of the student model at the $t$-th epoch. $\Theta_t$ denotes the parameters of the aggregated model. Since we can not foreknow the segmentation error on the target domain, we simply adopt average weights, which can be online updated as:
\begin{equation}
    \Theta_t = \frac{t-1}{t}\Theta_{t-1} + \frac{1}{t} \theta_t,
\label{eq:online}
\end{equation}
where $t\in\{2,...T_1\}$ and $\Theta_1 = \theta_1$. We do not need to update the mean student model very often. In our practice, we update the mean model every $T_2 = 5000$ iterations. \textbf{$T_1$ is the total training epoch number, while $T_2$ is the iteration number of each epoch.}
We find that the weight averaging still can achieve a competitive result. 

\noindent\textbf{Discussion.} 
\textbf{1. How about computation costs?} 
\emph{Time Cost.} As shown in Eq.~\ref{eq:online}, only 2 DNN weights need to be considered during every update, and the weight averaging costs about one forward inference time, which can be neglected~\cite{izmailov2018averaging}. \emph{Memory Cost.} The main memory cost is to preserve one mean student model. We note that the mean student model can be moved to CPU memory during idle time. We do not update the mean student model very often. When updating, we just move it back to GPU memory. Since we do not need to calculate the gradient but the mean of model parameters, the update process also costs limited memory.  
\textbf{2. What are the advantages of Student Aggregation?} The advantages are mostly in two aspects: First, instead of training $S~ (S>5)$ classifiers in the conventional AdaBoost, we only need to keep one mean student model, which can be preserved in CPU memory. Second, the model update is efficient and effective, which does not require gradient calculation. We only need to update the weights of every layer to the mean value of existing snapshots. Meanwhile, student aggregation provides one final model with good generalizability, preventing over-fitting as the traditional AdaBoost algorithm. More quantitative results are discussed in Experiment.
\textbf{3. Correlation with AdaBoost.} The main spirit of the proposed method is following Adaboost~\cite{freund1996experiments} to treat ``hard'' samples differently according to learned model snapshots. However, for domain adaptation, we can not foreknow the error $e_m$ on the unlabeled target-domain data. Therefore, we adopt the trade-off. 1) In this work,  we update the distribution based on the prediction variance instead of the error; 2) The binary classifier aggregation in Adaboost can be formulated as: $G(x)=sign(\sum_{m=1}^M \alpha_mG_m(x))$, where $G_m()$ is one of weak classifiers and $\alpha_m$ is the adaptive weights based on the error $e_m$. $\alpha_m = \frac{1}{2}log(\frac{1-e_m}{e_m})$. Since the error $e_m$ in target domain is unavailable, we deploy the temporal ensemble via weight averaging as a trade-off, which equals to set $\alpha_m =\frac{1}{M}$. In ablation studies, we also try different updating strategies, such as momentum, which is inferior to the proposed method. It verifies that the weight averaging trade-off is sub-optimal yet effective.

\begin{algorithm}[t]
\small
\caption{Training Procedure of the Proposed Method}
\label{alg:ada}
\begin{algorithmic}[1]
\Require The source domain dataset $X_m=\{X_m^i\}^M_{i=1}$; The source domain label $Y_m=\{y_m^i\}^M_{i=1}$; The unlabeled target domain dataset $X_n=\{x_n^j\}^N_{j=1}$; 
\Require The source-domain parameter $\theta_s$; The epoch number $T_1$; The iteration number $T_2$ in each epoch.
\State Initialize $\theta = \theta_s$;
\State Initialize target domain data distribution $D_1(j)=1/N$; 
\For {$epoch~t = 1$ to $T_1$}
\For {$iteration = 1$ to $T_2$}
%\State Input $x_n^j$ to $F(\cdot|\theta_t)$, extract the prediction of two classifiers, calculate the prediction variance according to Equation \ref{eq:var_final}:
%%%\vspace{-1.5ex}
%\begin{equation}
%D_{kl} = \E [F(x_n^j|\theta_t) \log (\frac{F(x_n^j|\theta_t)}{F_{aux}(x_n^j|\theta_t)})].
%\end{equation}
%%%\vspace{-2ex}
\State  To be simplify, here we only show the basic cross-entropy segmentation loss on the source domain data: 
%\vspace{-1.5ex}
\begin{equation}
L_{ce} = \E [- p_m^i \log F(x^i_m|\theta) ].
\end{equation}
\State We denotes other regularization losses on the target domain as $R(x^j_n, \theta)$. Update the model weight:
\begin{equation}
L_{total} = L_{ce} + R(x^j_n, \theta).
\end{equation}
%the prediction variance with the conventional objective to obtain the rectified objective. Update the $\theta_t$ according to Equation \ref{eq:rectified}:
%%\vspace{-1.5ex}
%\begin{equation}
%L_{rect} =  \E [  exp\{-D_{kl}\} L_{ce} +  D_{kl} %\end{equation}
\EndFor 
\State $\theta_t = \theta$.
\State The mean model parameter is updated:
\begin{equation}
\Theta_t = \frac{t-1}{t}\Theta_{t-1} + \frac{1}{t} \theta_t.
\label{eq:alg-student}
\end{equation}
\State Update the adaptive data distribution $D_{t+1}$:
\begin{equation}
D_{t+1}(j) = \frac{1}{2}(D_{t}(j) + V_{kl}(x_n^j|\Theta_t)),
\label{eq:alg-data}
\end{equation}
where $V_{kl}$ is the prediction variance defined in Eq.~\ref{eq:variance}, and we apply the current mean model $\Theta_t$ to estimate the variance.  
\EndFor \\
\Return $ \Theta_t $.
\end{algorithmic}
\end{algorithm}

\subsection{Optimization}
We combine the two components, \ie, adaptive sampler and student aggregation, in a coherent manner, and explicitly enable interactions between learned models and the data sampler. As shown in Algorithm~\ref{alg:ada}, the adaptive sampler depends on the mean student model $\Theta_t$ to update the data distribution (see Eq.~\ref{eq:alg-data}). Meanwhile, the adaptive sampler controls the input training data, which, in turn, affects the $\theta_t$ and $\Theta_t$ (see Eq.~\ref{eq:alg-student}).  
Since we do not pursue new segmentation losses or regularization objectives, here we just mention the basic source-domain segmentation loss as: 
\begin{equation}
    L_{ce} = \E [- p_m^i \log F(x^i_m|\theta) ].
\end{equation}
where $p_m^i$ is the ground-truth probability vector of the label $y_m^i$. The value $p_m^i(c)$ equals to $1$ if $c==y_m^i$ otherwise $0$. Besides, there are multiple feasible regularization terms, including adversarial losses proposed in~\cite{tsai2018learning,luo2019significance} and consistency loss~\cite{zheng2019unsupervised}. Without loss of generality, here we denote the regularization terms on the target domain as $R(x^j_n, \theta)$. Therefore, the optimization objective in every iteration can be formulated as: 
\begin{equation}
   L_{total} = L_{ce} + R(x^j_n, \theta).
\end{equation}
It is worth noting that different baseline methods have different regularization terms. We follow the regularization term in the corresponding baseline method for a fair comparison. In this paper, we compare three baseline methods, \ie, AdaptSegNet~\cite{tsai2018learning}, MRNet~\cite{zheng2019unsupervised} and Uncertainty~\cite{zheng2020unsupervised}.   AdaptSegNet mainly adopts the adversarial loss for regularization, while MRNet adds a consistency-based  memory regularization. The Uncertainty follows MRNet, so Uncertainty has the same regularization term as MRNet. For a fair comparison, we follow the corresponding regularization term in different baseline methods. After every $T_2$ iterations, we update the adaptive sampler and the mean student model, as discussed in Section~\ref{sec:sampler} and Section~\ref{sec:aggregation}. The training is stopped after $T_1$ epochs.

\section{Experiment}~\label{sec:experiment}
\subsection{Implementation Details}
\noindent\textbf{Network Architecture.} We adopt the Deeplab-v2~\cite{chen2017deeplab} as the baseline model, which deploys the ResNet-101~\cite{he2016deep} as the backbone. Following most existing works~\cite{tsai2018learning,tsai2019domain,luo2019significance,luo2019taking,zheng2019unsupervised,zheng2020unsupervised}, we insert one auxiliary classifier with the same structure as the primary classifier. The classifier consists of one Atrous Spatial Pyramid Pooling (ASPP) module~\cite{chen2017deeplab} and one fully-connected layer. The auxiliary classifier is added at the $res4b22$ layer. We also insert the dropout layer of $0.2$ drop rate before the fully-connected layer. %Besides, in the ablation study, we also study the scalability of our method on the relatively vanilla network, \ie, VGG16~\cite{simonyan2014very}. 

\noindent\textbf{Training Details.}
Following existing works~\cite{tsai2018learning, luo2019significance}, the input image is resized to $1280 \times 640$, and we randomly crop $1024 \times 512$ for training. Random horizontal flipping is also applied. We deploy the SGD optimizer with a mini-batch of 2. The initial learning rate is set as $0.0002$. Following~\cite{zhao2017pyramid,zhang2019dual,zhang2020dynamic}, we deploy the ploy learning rate policy by multiplying the scale factor $(1-\frac{current\_iter}{total\_iter})^{0.9}$. The total iteration is set as 50k ($T_1=10, T_2=5000$). 
When inference, we follow \cite{zheng2019unsupervised,zheng2020unsupervised} to combine the prediction of the two classifiers as the final prediction. Our implementation is based on Pytorch~\cite{paszke2017automatic}.

\subsection{Datasets and Evaluation Metric}
\noindent\textbf{Datasets.} In this work, we conduct experiments on three scene segmentation domain adaptation benchmarks. To simplify the illustration, we denote the source dataset $A$ and the target dataset $B$ as $A \rightarrow B$. 
Two widely-used benchmarks are to leverage the synthetic data to learn the shared knowledge, and adapt the knowledge to the real-world scenario. The settings are GTA5~\cite{richter2016playing} $\rightarrow$ Cityscapes~\cite{cordts2016cityscapes} and SYNTHIA~\cite{ros2016synthia} $\rightarrow$ Cityscapes~\cite{cordts2016cityscapes}. In particular, the GTA5 dataset is collected from one video game, which contains 24,966 labeled images for training. The SYNTHIA dataset is generated from a virtual city engine with pixel-level annotations as well, including 9,400 training images. The target dataset, \ie, Cityscapes, collects the real-world street-view images from $50$ different cities, yielding $2,975$ unlabeled training images and $500$ images for testing. 
We also adopt one cross-city benchmark, \ie, Cityscapes~\cite{cordts2016cityscapes} $\rightarrow$ Oxford RobotCar~\cite{RobotCarDatasetIJRR}. In this setting, we use the annotation of $2,975$ training images of the Cityscapes dataset. In contrast with the Synthetic-to-real setting, the Oxford RobotCar dataset is also collected from the realistic street-view camera. The challenge is in the noisy variants, such as weather and illumination conditions. The Oxford RobotCar dataset is collected on rainy days and cloudy days, while the Cityscapes dataset mostly contains images on sunny days. More details are shown in Table~\ref{table:Dataset}. 

\setlength{\tabcolsep}{5pt}
\begin{table}
\caption{List of categories and number of images of four datasets, \ie, GTA5~\cite{richter2016playing}, SYNTHIA~\cite{ros2016synthia}, Cityscapes~\cite{cordts2016cityscapes} and Oxford RobotCar~\cite{RobotCarDatasetIJRR}.
}
\label{table:Dataset}
\vspace{-.2in}
%\footnotesize
\begin{center}
\begin{tabular}{l|c|c|c|c}
\shline
Datasets & GTA5 & SYNTHIA & Cityscapes & Oxford RobotCar\\
\hline
\#Train Images & 24,966 & 9,400 & 2,975 & 894\\
\#Test Images & - & - & 500 & 271 \\
\#Category & 19 & 16 & 19 & 9 \\
Synthetic & \checkmark & \checkmark & $\times$ & $\times$ \\
\shline
\end{tabular}
\end{center}
\vspace{-.1in}
\end{table}

\noindent\textbf{Evaluation Metric.} We follow previous works to report the IoU accuracy of each category, and mean IoU (mIoU) for all the classes. For SYNTHIA $\rightarrow$ Cityscapes, some previous works report the mean IoU over $13$ classes, while others report both $13$ classes and $16$ classes. We report both results, and denote the results of $13$ classes and $16$ classes as mIoU$^*$ and mIoU, respectively. For GTA5 $\rightarrow$ Cityscapes and Cityscapes $\rightarrow$ Oxford RobotCar, we report the $19$-category and $9$-category accuracy, respectively. 

\subsection{Ablation Studies}

\begin{table}[tbp]
%\scriptsize
%\small
\caption{Ablation study of the two components in Adaboost Student on GTA5$\rightarrow$Cityscapes. We adopt MRNet~\cite{zheng2019unsupervised} as baseline, and study the impact of the adaptive sampler and the student aggregation alone. The results suggest that the adaptive sampler does not improve the single model performance but helps complementary model learning, facilitating the final ``weak'' student model aggregation. Ours (entropy) denotes that we apply the prediction entropy to update the adaptive sampler, while ours (variance) is the default setting if not specified. PutBack Sampler denotes random sampling with replacement.
} \label{table:ablation}
\vspace{-.2in}
%\scriptsize
\begin{center}
\resizebox{\linewidth}{!}
{
\setlength{\tabcolsep}{3pt}
\begin{tabular}{l|c|c|c|c}
\shline
Method  & PutBack  &  Adaptive & Student & mIoU \\
  & Sampler & Sampler & Aggregation & \\
\hline
MRNet~\cite{zheng2019unsupervised} $w$ Uniform & & & & 45.5 \\
$w$ PutBack  & $\checkmark$ &  & & 43.6 \\
$w$ Random & & & & 44.6 \\
$w$ Adaptive Sampler &  & $\checkmark$ & & 45.2 \\
$w$ Uniform + Student Aggregation &  & & $\checkmark$ & 48.4 \\
$w$ PutBack + Student Aggregation & $\checkmark$ &  & $\checkmark$ & 48.3 \\
$w$ Random + Student Aggregation & & & $\checkmark$ & 48.4 \\
MRNet~\cite{zheng2019unsupervised} + Ours (entropy)  & & $\checkmark$ & $\checkmark$ &  48.1 \\
MRNet~\cite{zheng2019unsupervised} + Ours (variance)  & & $\checkmark$ & $\checkmark$ & 49.0 \\
\shline
\end{tabular}}
\end{center}
%\vspace{-.1in}
\end{table}

\noindent\textbf{Effect of the Adaptive Sampler.}
First, we study the adaptive sampler. Since we intend to learn the complementary model snapshots, as shown in Table \ref{table:ablation}, the single ``weak'' model does not achieve significant improvement, even with the performance drop. The observation is aligned with the conventional AdaBoost algorithm~\cite{freund1996experiments}, since the current snapshot takes more attention to ``hard'' samples. Here we give one toy example on a binary classification to illustrate the phenomenon~\footnote{\url{https://github.com/layumi/AdaBoost_Seg/blob/master/Toy_Example.md}}.  It is because every weak classifier considers more hard negatives in the last round and may overfit the negative data. Despite the probability of overfitting the negatives, such snapshots are complementary to the snapshot trained in the early stage and help the ensembled model keep performance improvement. Therefore, compared with the baseline method, \ie, MRNet~\cite{zheng2019unsupervised}, which achieves $45.5\%$ mIoU accuracy, MRNet using adaptive sampler decreases to $45.2\%$. It is because the hard negatives do not lead to one single optimal student model but the complementary model of the last model snapshot. If we further apply the student aggregation, MRNet + Ours has achieved significant improvement from $45.2\%$ mIoU accuracy to $49.0\%$ mIoU accuracy, which also surpasses MRNet ($45.5\%$ mIoU) by a relatively large margin. By default, we apply the sampler based on the prediction variance, and we observe that the alternative sampler based on the prediction entropy leads to one inferior performance of $48.1\%$ mIoU. It is mainly due to the prediction entropy taking more attention to the object edges as shown in~\cite{zheng2020unsupervised}, which can not nominate ``hard'' samples with ambiguous predictions accurately. %rather than ``hard'' samples with disagreement between classifiers.

Furthermore, we compare other sampling strategies, \ie, sampling with replacement (PutBack) and random sampling (Random), to verify the effectiveness of the proposed Adaptive Sampler. PutBack sampler is different from the original MRNet, which directly adopts the sampling without replacement (Uniform). 
As shown in Table~\ref{table:ablation}, (1) PutBack Sampler does not lead to a better single snapshot. It is because the learned model does not ``see’’ all data in every iteration, which compromises the training process. (2) PutBack Sampler also ensures diverse snapshots. Therefore, PutBack Sampler + Student Aggregation arrives $48.3\%$ mIoU accuracy, which is close to the baseline +  Student Aggregation of $48.4\%$. (3) The proposed Adaptive Sampler (ours) surpasses the PutBack Sampler, since we explicitly add more emphasis on the hard samples. Similarly, we could observe that (1) Random sampling (44.6\%) is inferior to Uniform sampling (45.5\%). It is because the random sampler, like PutBack sampler, usually does not ``see’’ all data in every epoch. (2) Random sampling leads to diverse snapshots, but it also suffers from a larger performance fluctuation.  We observe that some snapshots in our implementation arrive at around 41-42\% mIoU, which results in performance convergence.  For instance, the snapshot at the 25000th iteration performs 44.6\% mIoU, but the model at the 45000th iteration only arrives at 41.8\% mIoU. Therefore, Random sampling + student aggregation (averaged model) only converges to 48.4\% mIoU, which is also inferior to our method. Due to the random sampling is not stable, we run the experiment twice. Despite the performance fluctuation of the single snapshot, the re-run performance with random sampling also converges to  48.3\% mIoU (around 48.4\% mIoU). Finally, we could see that no matter uniform, putback, or random sampling method, the aggregated models all converge to around 48.4\% compared to the 49.0\% of our method. Therefore, it verifies our motivation that the sampling on hard negative samples is non-trivial. 

\noindent\textbf{Effect of the Student Aggregation.} 
We also investigate the student aggregation in Adaboost Student (see Table \ref{table:ablation}). There are two main observations. (1) The student aggregation alone can yield performance improvement. It is due to the model discrepancy during training. Especially for the scene segmentation, the extremely small training batch sizes and data shuffle lead to different training orders in every epoch. Therefore, the baseline method, \ie, MRNet, with student aggregation alone, has arrived at $48.4\%$ mIoU accuracy from $45.5\%$ mIoU. 
(2) Since the input data distribution is fixed in the baseline method, the model snapshots are still homogeneous due to the training data. In contrast, the proposed method explicitly demands learning complementary models by the dynamic input data via the adaptive sampler. The full method with the adaptive sampler, therefore, yields better results at $49.0\%$, surpassing the student aggregation alone ($48.4\%$). The results also suggest the effectiveness of the adaptive sampler.

\begin{figure}[t]
\begin{center}
     \includegraphics[width=1\linewidth]{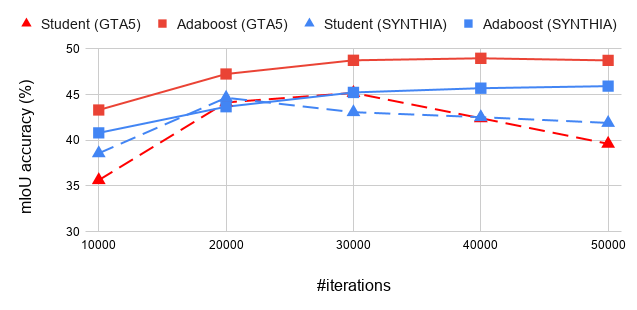}
\end{center} 
\vspace{-.2in}
      \caption{ Convergence of the proposed method. We provide the sensitivity analysis of MRNet+Ours on two benchmarks, \ie, GTA5$\rightarrow$Cityscapes (GTA5) and SYNTHIA$\rightarrow$Cityscapes (SYNTHIA). The results suggest that the AdaBoost Student model (AdaBoost) quickly converges, while the performance of the student model (Student) is still fluctuating. }
      \label{fig:convergence}
\end{figure}

\begin{table*}[!t]
	\centering
	\caption{Quantitative results on GTA5 $\rightarrow$ Cityscapes. We present pre-class IoU and mIoU. The best accuracy in every column is in \textbf{bold}.}
	\label{table:gtacity}
	\resizebox{\linewidth}{!}{
	\begin{tabular}{c|ccccccccccccccccccc|c}
		\shline
		Method & Road & SW & Build & Wall & Fence & Pole & TL & TS & Veg. & Terrain & Sky & PR & Rider & Car & Truck & Bus & Train & Motor & Bike & mIoU\\
		\shline
		%Source & \multirow{2}{0.1\linewidth}{\centering{DRN-26}} & 42.7 & 26.3 & 51.7 & 5.5 & 6.8 & 13.8 & 23.6 & 6.9  & 75.5 & 11.5 & 36.8 & 49.3 & 0.9 & 46.7 & 3.4 & 5.0 & 0.0 & 5.0 & 1.4  & 21.7\\
		%CyCADA~\cite{hoffman2018cycada} & & 79.1 & 33.1 & 77.9 & 23.4 & 17.3 & 32.1 & 33.3 & 31.8 & 81.5 & 26.7 & 69.0 & 62.8 & 14.7 & 74.5 & 20.9 & 25.6 & 6.9 & 18.8 & 20.4 & 39.5\\
		%\hline
		%Source & \multirow{2}{0.1\linewidth}{\centering{DRN-105}} & 36.4 & 14.2 & 67.4 & 16.4 & 12.0 & 20.1 & 8.7 & 0.7 & 69.8 & 13.3 & 56.9 & 37.0 & 0.4 & 53.6 & 10.6 & 3.2 & 0.2 & 0.9 & 0.0 & 22.2\\
		%MCD~\cite{saito2018maximum} & & 90.3 & 31.0 & 78.5 & 19.7 & 17.3 & 28.6 & 30.9 & 16.1 & 83.7 & 30.0 & 69.1 & 58.5 & 19.6 & 81.5 & 23.8 & 30.0 & 5.7 & 25.7 & 14.3 & 39.7\\
		\hline
		Source & 75.8 & 16.8 & 77.2 & 12.5 & 21.0 & 25.5 & 30.1 & 20.1 & 81.3 & 24.6 & 70.3 & 53.8 & 26.4 & 49.9 & 17.2 & 25.9 & 6.5 & 25.3 & 36.0 & 36.6\\
		AdaptSegNet~\cite{tsai2018learning} & 86.5 & 36.0 & 79.9& 23.4 & 23.3 & 23.9 & 35.2 & 14.8 & 83.4 & 33.3 & 75.6 & 58.5 & 27.6 & 73.7 & 32.5 & 35.4 & 3.9 & 30.1 & 28.1 & 42.4\\ 
		\rowcolor{lightgray} 
		AdaptSegNet~\cite{tsai2018learning} + Ours & 89.5 & 34.1 & 83.7 & 31.5 & 25.0 & 36.0 & 40.6 & 34.6 & 85.6 & \textbf{44.6} & 77.8 & 61.4 & 30.4 & 85.6 & 35.9 & 47.1 & 1.8 & 19.4 & 34.1 & 47.3
		\\
		SIBAN \cite{luo2019significance} & 88.5 & 35.4 & 79.5 & 26.3 & 24.3 & 28.5 & 32.5 & 18.3 & 81.2 & 40.0 & 76.5 & 58.1 & 25.8 & 82.6 & 30.3 & 34.4 & 3.4 & 21.6 & 21.5 & 42.6 \\
		CLAN~\cite{luo2019taking} & 87.0 & 27.1 & 79.6 & 27.3 & 23.3 & 28.3 & 35.5 & 24.2 & 83.6 & 27.4 & 74.2 & 58.6 & 28.0 & 76.2 & 33.1 & 36.7 & 6.7 & 31.9 & 31.4 & 43.2 \\
		SP-Adv~\cite{SHAN2020125} & 86.2 & 38.4 & 80.8 & 25.5& 20.5 & 32.8 & 33.4 & 28.2 & 85.5 & 36.1 & 80.2 & 60.3 & 28.6 & 78.7 & 27.3 & 36.1 &	4.6	& 31.6 & 28.4 & 44.3 \\
		MaxSquare~\cite{chen2019domain} & 88.1 & 27.7 & 80.8 & 28.7 & 19.8 & 24.9 & 34.0 & 17.8 & 83.6 & 34.7 & 76.0 & 58.6 & 28.6 & 84.1 & 37.8 & 43.1 & 7.2 & 32.3 & 34.2 & 44.3 \\
		ASA~\cite{zhou2020affinity} & 89.2 & 27.8 & 81.3 & 25.3 & 22.7 & 28.7 & 36.5 & 19.6 & 83.8 & 31.4 & 77.1 & 59.2 & 29.8 & 84.3 & 33.2 & 45.6 & 16.9 & \textbf{34.5} & 30.8 & 45.1 \\
		APODA~\cite{yang2020adversarial} & 85.6 & 32.8 & 79.0 & 29.5 & 25.5 & 26.8 & 34.6 & 19.9 & 83.7 & 40.6 & 77.9 & 59.2 & 28.3 & 84.6 & 34.6 & 49.2 & 8.0 & 32.6 & 39.6 & 45.9 \\
		PatchAlign~\cite{tsai2019domain} & \textbf{92.3} & 51.9 & 82.1 & 29.2 & 25.1 & 24.5 & 33.8 & 33.0 & 82.4 & 32.8 & 82.2 & 58.6 & 27.2 & 84.3 & 33.4 & 46.3 & 2.2 & 29.5 & 32.3 & 46.5 \\
		BL~\cite{li2019bidirectional} & 91.0 & 44.7 & 84.2 & 34.6 & 27.6 & 30.2 & 36.0 & 36.0 & 85.0 & 43.6 & 83.0 & 58.6 & 31.6 & 83.3 & 35.3 & 49.7 & 3.3 & 28.8 & 35.6 & 48.5 \\
		DT~\cite{wang2020differential} & 90.6 & 44.7 & 84.8 & 34.3 & 28.7 & 31.6 & 35.0 & 37.6& 84.7 & 43.3 & \textbf{85.3} & 57.0 & 31.5 & 83.8 & \textbf{42.6} & 48.5 & 1.9 & 30.4 & 39.0 & 49.2\\
		\hline
		AdvEnt~\cite{vu2019advent} & 89.4 & 33.1 & 81.0 & 26.6 & 26.8 & 27.2 & 33.5 & 24.7 & 83.9 & 36.7 & 78.8 & 58.7 & 30.5 & 84.8 & 38.5 & 44.5 & 1.7 & 31.6 & 32.4 & 45.5 \\
		\hline
		Source &  - & - & - & - & - & - & - & - & -& - & - & - & - & - & - & - & - & - & - & 29.2\\
		FCAN~\cite{zhang2018fully} & - & - & - & - & - & - & - & - & -& - & - & - & - & - & - & - & - & - & - & 46.6 \\
		\hline
		Source & 71.3 & 19.2 & 69.1 & 18.4 & 10.0 & 35.7 & 27.3 &  6.8 & 79.6 & 24.8 & 72.1 & 57.6 & 19.5 & 55.5 & 15.5 & 15.1 & 11.7 & 21.1 & 12.0 & 33.8\\
		CBST \cite{zou2018unsupervised} & 91.8 & 53.5 & 80.5 & 32.7 & 21.0 & 34.0 & 28.9 & 20.4 & 83.9 & 34.2 & 80.9 & 53.1 & 24.0 & 82.7 & 30.3 & 35.9 & 16.0 & 25.9 & 42.8 & 45.9\\
		%MRL2 & & \textbf{91.9} & 55.2 & 80.9 & 32.1 & 21.5 & 36.7 & 30.0 & 19.0 & 84.8 & 34.9 & 80.1 & 56.1 & 23.8 & 83.9 & 28.0 & 29.4 & 20.5 & 24.0 & 40.3 & 46.0\\
		%MRENT & & 91.8 & 53.4 & 80.6 & 32.6 & 20.8 & 34.3 & 29.7 & 21.0 & 84.0 & 34.1 & 80.6 & 53.9 & 24.6 & 82.8 & 30.8 & 34.9 & 16.6 & 26.4 & 42.6 & 46.1\\
		MRKLD \cite{zou2019confidence} & 91.0 & \textbf{55.4} & 80.0 & 33.7 & 21.4 & 37.3 & 32.9 & 24.5 & 85.0 & 34.1 & 80.8 & 57.7 & 24.6 & 84.1 & 27.8 & 30.1 & \textbf{26.9} & 26.0 & 42.3 & 47.1\\
		%LRENT & & 91.8 & 53.5 & 80.5 & 32.7 & 21.0 & 34.0 & 29.0 & 20.3 & 83.9 & 34.2 & \textbf{80.9} & 53.1 & 23.9 & 82.7 & 30.2 & 35.6 & 16.3 & 25.9 & 42.8 & 45.9\\
		\hline
		%Source &\multirow{3}{0.1\linewidth}{\centering{DeepLabv2}} & 51.13 & 18.33 & 75.8 & 18.81 & 16.81 & 34.69 & 36.26 & 27.18 & 80.0 & 23.25 & 64.93 & 59.15 & 19.29 & 74.62 & 26.71 & 13.77 & 0.11 & 32.44 & 34.03 & 37.23\\
		%Our (Stage-I)  & & 89.08 & 23.88 & 82.18 & 19.51 & 20.14 & 33.54 & 42.23 & 39.05 & \textbf{85.32} & 33.65 & 76.4 & 60.19 & 33.66 & \textbf{85.97} & 36.13 & 43.27 & 5.9 & 22.83 & 30.78 & 45.46 \\
		Source & 51.1 & 18.3 & 75.8 & 18.8 & 16.8 & 34.7 & 36.3 & 27.2 & 80.0 & 23.3 & 64.9 & 59.2 & 19.3 & 74.6 & 26.7 & 13.8 & 0.1 & 32.4 & 34.0 & 37.2\\
		MRNet~\cite{zheng2019unsupervised}  & 89.1 & 23.9 & 82.2 & 19.5 & 20.1 & 33.5 & 42.2 & 39.1 & 85.3 & 33.7 & 76.4 & 60.2 & 33.7 & 86.0 & 36.1 & 43.3 & 5.9 & 22.8 & 30.8 & 45.5 \\
		\rowcolor{lightgray} 
		MRNet~\cite{zheng2019unsupervised} + Ours & 
		89.5 & 36.6 & 84.8 & 37.9 & \textbf{29.8} & 
		37.9 & 45.1 & 34.9 & 85.8 & 43.6 &
		80.4 & \textbf{64.7} & 35.3 & 84.9 & 30.4 & 
		45.9 & 2.7 & 24.7 & 35.7 & 49.0\\
		%Our (Stage-II) & & 90.51 & 34.99 & \textbf{84.64} & \textbf{34.3} & 23.96 & 36.76 & \textbf{44.09} & \textbf{42.74} & 84.47 & 33.55 & \textbf{82.51} & \textbf{63.08} & \textbf{34.35} & 85.84 & 32.93 & 38.18 & 2.0 & 27.1 & 41.79 & \textbf{48.31} \\
		%MRNet+Pseudo~\cite{zheng2019unsupervised} & 90.5 & 35.0 & 84.6 & 34.3 & 24.0 & 36.8 & 44.1 & 42.7 & 84.5 & 33.6 & \textbf{82.5} & 63.1 & 34.4 & 85.8 & 32.9 & 38.2 & 2.0 & 27.1 & 41.8 & 48.3 \\
		Uncertainty~\cite{zheng2020unsupervised} & 90.4 & 31.2 & 85.1 & 36.9 & 25.6 & 37.5 & 48.8 & \textbf{48.5} & 85.3 & 34.8 & 81.1 & 64.4 & \textbf{36.8} & 86.3 & 34.9 & \textbf{52.2} & 1.7 & 29.0 & 44.6 & 50.3 \\
		\rowcolor{lightgray} 
		Uncertainty~\cite{zheng2020unsupervised} + Ours & 90.7 & 35.9 & \textbf{85.7} & \textbf{40.1} & 27.8 & \textbf{39.0} & \textbf{49.0} & 48.4 & \textbf{85.9} & 35.1 & 85.1 & 63.1 & 34.4 & \textbf{86.8} & 38.3 & 49.5 & 0.2 & 26.5 & \textbf{45.3} & \textbf{50.9}\\
		\shline
	\end{tabular}
	}
\end{table*}

\begin{table*}[!t]
	\centering
	\caption{Quantitative results on SYNTHIA $\rightarrow$ Cityscapes. We present pre-class IoU, mIoU and mIoU*. mIoU and mIoU* are averaged over 16 and 13 categories, respectively. The best accuracy in every column is in \textbf{bold}.}
	\label{table:syncity}
	\resizebox{\linewidth}{!}{
	\begin{tabular}{c|cccccccccccccccc|c|c}
		\shline
		Method & Road & SW & Build & Wall* & Fence* & Pole* & TL & TS & Veg. & Sky & PR & Rider & Car & Bus & Motor & Bike & mIoU* & mIoU\\
		\shline
		%Source & \multirow{2}{0.1\linewidth}{\centering{DRN-105}} & 14.9 & 11.4 & 58.7 & 1.9 & 0.0 & 24.1 & 1.2 & 6.0 & 68.8 & 76.0 & 54.3 & 7.1 & 34.2 & 15.0 & 0.8 & 0.0 & 26.8 & 23.4\\
		%MCD \cite{saito2018maximum} & & 84.8 & \textbf{43.6} & 79.0 & 3.9 & 0.2 & 29.1 & 7.2 & 5.5 & 83.8 & 83.1 & 51.0 & 11.7 & 79.9 & 27.2 & 6.2 & 0.0 & 43.5 & 37.3 \\
		\hline
		Source & 55.6 & 23.8 & 74.6 & $-$ & $-$ & $-$ & 6.1 & 12.1 & 74.8 & 79.0 & 55.3 & 19.1 & 39.6 & 23.3 & 13.7 & 25.0 & 38.6 & $-$ \\
		MaxSquare~\cite{chen2019domain} & 77.4 & 34.0 & 78.7  & 5.6 & 0.2 & 27.7 & 5.8 & 9.8 & 80.7 & 83.2 & 58.5 & 20.5 & 74.1 & 32.1 & 11.0 & 29.9 & 45.8 & 39.3 \\
		SIBAN~\cite{luo2019significance} & 82.5 & 24.0 & 79.4 & $-$ & $-$ & $-$ & 16.5 & 12.7 & 79.2 & 82.8 & 58.3 & 18.0 & 79.3 & 25.3 & 17.6 & 25.9 & 46.3 & $-$ \\
    	PatchAlign~\cite{tsai2019domain} & 82.4 & 38.0 & 78.6 & 8.7 & 0.6 & 26.0 & 3.9 & 11.1 & 75.5 & 84.6 & 53.5 & 21.6 & 71.4 & 32.6 & 19.3 & 31.7 & 46.5 & 40.0 \\
		AdaptSegNet~\cite{tsai2018learning} & 84.3 & 42.7 & 77.5 & $-$ & $-$ & $-$ & 4.7 & 7.0 & 77.9 & 82.5 & 54.3 & 21.0 & 72.3 & 32.2 & 18.9 & 32.3 & 46.7 & $-$ \\
		\rowcolor{lightgray} 
		AdaptSegNet~\cite{tsai2018learning} + Ours & 67.9 & 30.1 & 77.9 & 10.2 & 1.5 & 37.2 & 30.9 & 22.3 & 80.8 & 83.1 & 52.4 & 20.9 & 72.9 & 25.4 & 12.7 & 45.7 & 47.9 & 42.0  \\
		CLAN~\cite{luo2019taking} & 81.3 & 37.0 & 80.1 & $-$ & $-$ & $-$ & 16.1 & 13.7 & 78.2 & 81.5 & 53.4 & 21.2 & 73.0 & 32.9 & 22.6 & 30.7 & 47.8 & $-$  \\
		SP-Adv~\cite{SHAN2020125} & 84.8 & 	35.8 & 	78.6 & $-$ & $-$ & $-$ &	6.2	& 15.6 & 	80.5 &	82.0 &	66.5 & 	22.7 & 	74.3 &	34.1	 & 19.2	 & 27.3 & 	48.3 &  $-$ \\
		ASA~\cite{zhou2020affinity} & \textbf{91.2} & \textbf{48.5} & 80.4 & 3.7 & 0.3 & 21.7 & 5.5 & 5.2 & 79.5 & 83.6 & 56.4 & 21.0 & 80.3 & 36.2 & 20.0 & 32.9 & 49.3 & 41.7 \\
		DADA~\cite{vu2019dada} & 89.2 & 44.8 & 81.4 & 6.8 & 0.3 & 26.2 & 8.6 & 11.1 & 81.8 & 84.0 & 54.7 & 19.3 & 79.7 & 40.7 & 14.0 & 38.8 & 49.8 & 42.6 \\
		BL~\cite{li2019bidirectional} & 86.0 & 46.7 & 80.3 & $-$ & $-$ & $-$ & 14.1 & 11.6 & 79.2 & 81.3 & 54.1 & 27.9 & 73.7 & 42.2 & 25.7 & 45.3 & 51.4 & $-$ \\
		DT~\cite{wang2020differential} & 83.0 & 44.0 & 80.3 & $-$ & $-$ & $-$ & 17.1 & 15.8 & 80.5 & 81.8 & 59.9 & 33.1 & 70.2 & 37.3 & 28.5 & 45.8 & 52.1 &  $-$ \\
		CCM~\cite{li2020content} & 79.6 & 36.4 & 80.6 & 13.3 & 0.3 & 25.5 & 22.4 & 14.9 & 81.8 & 77.4 & 56.8 & 25.9 & 80.7 & 45.3 & 29.9 &  52.0 & 52.9 & 45.2 \\
		APODA~\cite{yang2020adversarial} & 86.4 & 41.3 & 79.3 & $-$ & $-$ & $-$ & 22.6 & 17.3 & 80.3 & 81.6 & 56.9 & 21.0 & 84.1 & \textbf{49.1} & 24.6 & 45.7 & 53.1 & $-$  \\
		\hline
		AdvEnt \cite{vu2019advent} & 85.6 & 42.2 & 79.7 & 8.7 & 0.4 & 25.9 & 5.4 & 8.1 & 80.4 & 84.1 & 57.9 & 23.8 & 73.3 & 36.4 & 14.2 & 33.0 & 48.0 & 41.2 \\
		\hline
		Source & 64.3 & 21.3 & 73.1 & 2.4 & 1.1 & 31.4 & 7.0 & 27.7 & 63.1 & 67.6 & 42.2 & 19.9 & 73.1 & 15.3 & 10.5 & 38.9 & 40.3 & 34.9 \\
		CBST \cite{zou2018unsupervised} & 68.0 & 29.9 & 76.3 & 10.8 & 1.4 & 33.9 & 22.8 & 29.5 & 77.6 & 78.3 & 60.6 & 28.3 & 81.6 & 23.5 & 18.8 & 39.8 & 48.9 & 42.6 \\
		MRKLD \cite{zou2019confidence} & 67.7 & 32.2 & 73.9 & 10.7 & 1.6 & 37.4 & 22.2 & \textbf{31.2} & 80.8 & 80.5 & 60.8 & \textbf{29.1} & 82.8 & 25.0 & 19.4 & 45.3 & 50.1 & 43.8 \\
		\hline
		%Source & \multirow{3}{0.1\linewidth}{\centering{DeepLabv2}} & 44.01 & 19.31 & 70.86 & 8.65 & 0.76 & 28.19 & 16.14 & 16.7 & 79.84 & 81.35 & 57.77 & 19.17 & 46.88 & 17.21 & 12.01 & 43.78 & 40.39 & 35.16 \\
		%Ours (Stage-I) & & 81.96 & 36.5 & 80.35 & 4.23 & 0.37 & 33.65 & 18.04 & 13.41 & 81.08 & 80.83 & 61.34 & 21.68 & 84.36 & 32.36 & 14.84 & 45.72 & 50.19 & 43.17 \\
		%Ours (Stage-II) & & 83.13 & 38.22 & \textbf{81.72} & 9.31 & 1.01 & 35.14 & \textbf{30.25} & 19.88 & \textbf{81.99} & 80.11 & \textbf{62.79} & 21.05 & \textbf{84.37} & 37.81 & 24.52 & \textbf{53.28} & \textbf{53.78} & \textbf{46.54} 
		Source & 44.0 & 19.3 & 70.9 & 8.7 & 0.8 & 28.2 & 16.1 & 16.7 & 79.8 & 81.4 & 57.8 & 19.2 & 46.9 & 17.2 & 12.0 & 43.8 & 40.4 & 35.2 \\
		MRNet~\cite{zheng2019unsupervised} & 82.0 & 36.5 & 80.4 & 4.2 & 0.4 & 33.7 & 18.0 & 13.4 & 81.1 & 80.8 & 61.3 & 21.7 & 84.4 & 32.4 & 14.8 & 45.7 & 50.2 & 43.2 \\
		\rowcolor{lightgray} 
		MRNet~\cite{zheng2019unsupervised} + Ours & 83.6 & 40.1 & 81.3 & 9.6 & 0.8 & 36.8 & 25.8 & 16.3 & 84.5 & 87.1 & 60.3 & 23.9 & 84.9 & 32.3 & 19.0 & 48.4 & 52.9 & 45.9 \\
		%MRNet+Pesudo~\cite{zheng2019unsupervised} & 83.1 & 38.2 & 81.7 & 9.3 & 1.0 & 35.1 & 30.3 & 19.9 & \textbf{82.0} & 80.1 & 62.8 & 21.1 & 84.4 & 37.8 & 24.5 & \textbf{53.3} & 53.8 & 46.5 \\
		Uncertainty~\cite{zheng2020unsupervised} & 87.6 & 41.9 & 83.1 & 14.7 & \textbf{1.7} & 36.2 & 31.3 & 19.9 & 81.6 & 80.6 & 63.0 & 21.8 & 86.2 & 40.7 & 23.6 & \textbf{53.1} & 54.9 & 47.9 
		\\
		\rowcolor{lightgray} 
		Uncertainty~\cite{zheng2020unsupervised} + Ours & 85.6 & 43.9 & \textbf{83.9} & \textbf{19.2} & \textbf{1.7} & \textbf{38.0} & \textbf{37.9} & 19.6 & \textbf{85.5} & \textbf{88.4} & \textbf{64.1} & 25.7 & \textbf{86.6} & 43.9 & \textbf{31.2} & 51.3 & \textbf{57.5} & \textbf{50.4} \\
		\shline
	\end{tabular}
	}
\end{table*}

\begin{table} [!t]
	%	\scriptsize
	\centering
	\caption{
		Quantitative results on the cross-city benchmark: Cityscapes $\rightarrow$ Oxford RobotCar. The best accuracy is in \textbf{bold}.
	}
		\label{table:oxford}
    \resizebox{\linewidth}{!}{
	\begin{tabular}{l|ccccccccc|c}
		\shline
		Method & \rotatebox{90}{road} & \rotatebox{90}{sidewalk} & \rotatebox{90}{building} & \rotatebox{90}{light} & \rotatebox{90}{sign} & \rotatebox{90}{sky} & \rotatebox{90}{person} & \rotatebox{90}{automobile} & \rotatebox{90}{two-wheel} & mIoU\\
		
		\shline
		
		Source & 79.2 & 49.3 & 73.1 & 55.6 & 37.3 & 36.1 & 54.0 & 81.3 & 49.7 & 61.9 \\
		AdaptSegNet~\cite{tsai2018learning}  & 95.1 & 64.0 & 75.7 & 61.3 & 35.5 & 63.9 & 58.1 & 84.6 & 57.0 & 69.5 \\
		\rowcolor{lightgray} 
		AdaptSegNet~\cite{tsai2018learning} + Ours & 96.0 & 73.0 & 91.5 & 65.8 & 22.9 & \textbf{94.9} & 56.0 & 89.0 & 53.0 & 71.3 \\
		PatchAlign~\cite{tsai2019domain} & 94.4 & 63.5 & 82.0 & 61.3 & 36.0 & 76.4 & 61.0 & 86.5 & 58.6 & 72.0 \\
		\hline
		%Source & 92.5 & 66.3 & 88.5 & 68.0 & 8.0 & 89.1 & 28.0 & 86.3 & 55.9 & 64.7 \\
		MRNet~\cite{zheng2019unsupervised} & 95.9 & 73.5 & 86.2 & 69.3 & 31.9 & 87.3 & 57.9 & 88.8 & \textbf{61.5} & 72.5 \\
		\rowcolor{lightgray} 
		MRNet~\cite{zheng2019unsupervised} + Ours & \textbf{96.4} & \textbf{77.0} & 83.4 & 70.7 & 39.2 & 83.0 & 62.4 & \textbf{89.9} &61.0 & 73.7 \\
		%MRNet+Pseudo~\cite{zheng2019unsupervised} & 95.1 & 72.5 & 87.0 & 72.2 & 37.4 & 87.9 & \textbf{63.4} & \textbf{90.5} & 58.9 & 73.9 \\
		Uncertainty~\cite{zheng2020unsupervised} & 95.9 & 73.7 & 87.4 & \textbf{72.8} & \textbf{43.1} & 88.6 & 61.7 & 89.6 & 57.0 & 74.4 \\
		\rowcolor{lightgray} 
		Uncertainty~\cite{zheng2020unsupervised} + Ours & 96.1 & 75.3 & \textbf{92.1} & 72.6 & 37.0 & 94.4 & \textbf{62.7} & 89.7 & 56.9 & \textbf{75.2} \\
		\shline
	\end{tabular}}
\end{table}

\noindent\textbf{Compatibility of the Proposed Method.} 
We argue that the proposed method is orthogonal to most existing methods, and can be fused with existing frameworks to obtain better performance. 
Therefore, we re-implement three existing approaches, including AdaptSegNet~\cite{tsai2018learning}, MRNet~\cite{zheng2019unsupervised} and Uncertainty~\cite{zheng2020unsupervised}. As shown in Table~\ref{table:gtacity}, we observe one consistent performance improvement on GTA5$\rightarrow$Cityscapes for all three methods. For the widely-adopted method, \ie, AdaptSegNet~\cite{tsai2018learning}, AdaptSegNet + Ours has arrived at $47.3\%$ mIoU, which surpasses the original AdaptSegNet ($42.4\%$) by $+4.9\%$. 
Furthermore, we observe one similar performance improvement on recent methods. MRNet~\cite{zheng2019unsupervised} + Ours has boosted from $45.5\%$ mIoU accuracy to $49.0\%$, while Uncertainty~\cite{zheng2020unsupervised} + Ours obtains $+0.6\%$ from $50.3\%$ to $50.9\%$ mIoU accuracy.   
As shown in Table~\ref{table:syncity} and Table~\ref{table:oxford}, we observe one similar result on both SYNTHIA$\rightarrow$Cityscapes and Cityscapes$\rightarrow$Oxford RobotCar benchmarks. For the SYNTHIA$\rightarrow$Cityscapes benchmark, AdaptSegNet~\cite{tsai2018learning}, MRNet~\cite{zheng2019unsupervised} and Uncertainty~\cite{zheng2020unsupervised} have achieved $+1.2\%$, $+2.7\%$ and $+2.6\%$ mIoU*, respectively.
On the other hand, AdaptSegNet~\cite{tsai2018learning}, MRNet~\cite{zheng2019unsupervised} and Uncertainty~\cite{zheng2020unsupervised} have achieved $+1.8\%$, $+1.2\%$ and $+0.8\%$ mIoU on the Cityscapes$\rightarrow$Oxford RobotCar benchmark, respectively.
%\zznote{As shown in Table~\ref{table:vgg}, the proposed method significantly improves MRNet with VGG16-based network from XX\% mIoU to XX\% mIoU, yielding a competitive performance with other methods.}
In a summary, the extensive experiment with three methods on three benchmarks verifies that the proposed approach, Adaptive Student, can be fused with existing frameworks to improve the model generalizability in the target domain.

\noindent\textbf{Limitation.} We observe that the proposed method shows less improvement on the pseudo label learning based method, \ie, Uncertainty~\cite{zheng2020unsupervised} than the domain alignment based method, \ie, AdaptSegNet~\cite{tsai2018learning} and MRNet~\cite{zheng2019unsupervised}. The main reason is the strong supervised learning of the generated pseudo labels on the target domain, which limits the model discrepancy and compromises learning complementary ``weak'' models in the proposed AdaBoost Student.

\noindent\textbf{Convergence of the Proposed Method.}
One advantage of the proposed method is the quick convergence to one robust model after limited iterations, since the complementary models are aggregated during training. Here we provide the sensitivity analysis of convergence time on both GTA5 $\rightarrow$ Cityscapes and SYNTHIA $\rightarrow$ Cityscapes. We adopt MRNet as the baseline model. As shown in Figure~\ref{fig:convergence}, we observe that the proposed method can efficiently converge at around $30,000$ iterations, while the student models are still jittering during training.

\begin{figure}[t]
\begin{center}
     \includegraphics[width=1\linewidth]{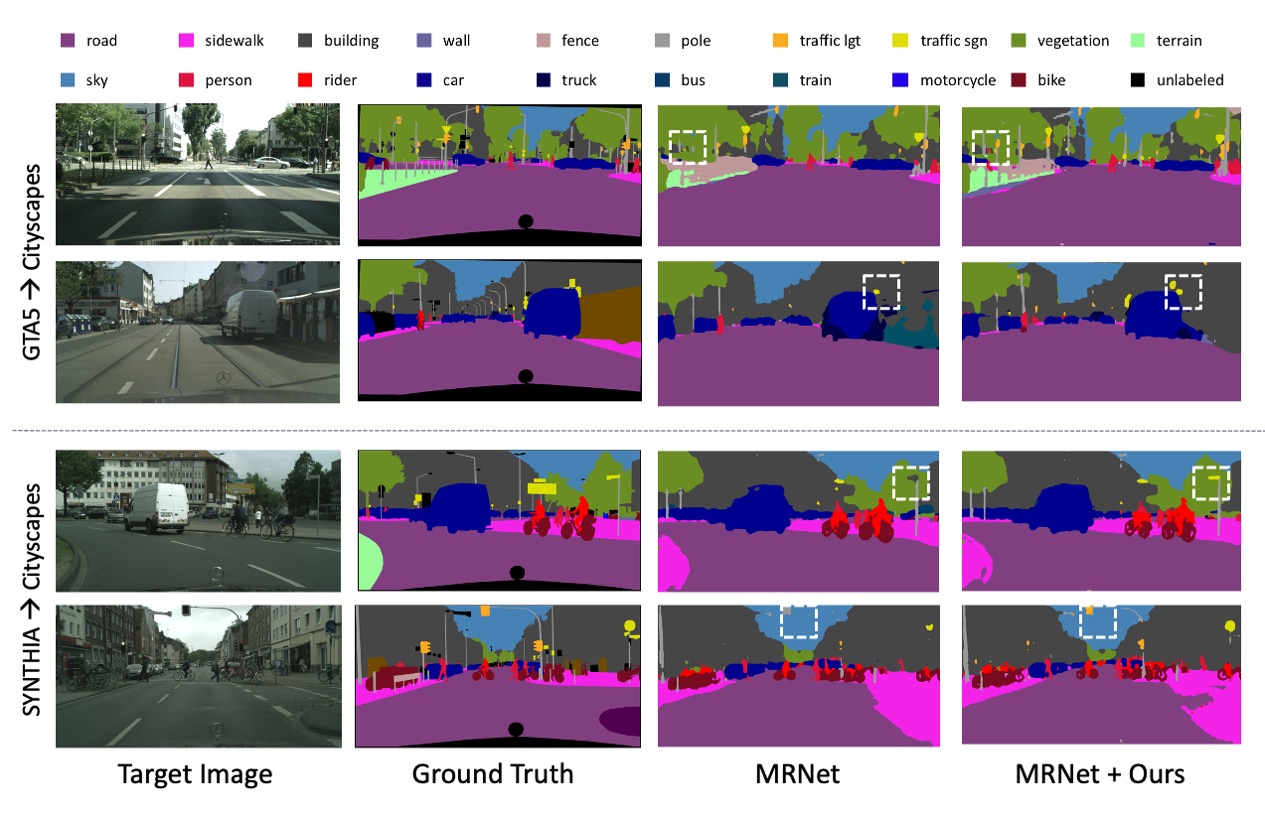}
\end{center} 
\vspace{-.3in}
      \caption{  Qualitative results of semantic segmentation adaptation on GTA5 $\rightarrow$ Cityscapes and SYNTHIA $\rightarrow$ Cityscapes. We show the original target-domain image, the ground-truth segmentation, and the output of methods, \ie, MRNet~\cite{zheng2019unsupervised}, and MRNet~\cite{zheng2019unsupervised} + Ours.  Our results are in the right column. (Best viewed in \emph{color}). We deploy the white dash boxes to highlight the different predictions. The proposed method is more robust to the minority categories, such as lights, poles, and traffic signs.}
      \label{fig:result}
\end{figure}

\subsection{Comparisons with State-of-the-art Methods} \noindent\textbf{Synthetic-to-real.} We mainly compare other recent methods on two synthetic-to-real benchmarks, which leverage the synthetic data to learn the shared knowledge. We compare recent methods with reported results and several methods re-implemented by us. 
For a fair comparison, we compare models with a similar structure, which is based on the Deeplab-v2~\cite{chen2017deeplab}, and the backbone is ResNet-101~\cite{he2016deep}. 
The competitive methods can be generally classified into two categories according to the usage of the pseudo label. One line of works focuses on the domain alignment, including AdaptSegNet~\cite{tsai2018learning}, SIBAN~\cite{luo2019significance}, CLAN~\cite{luo2019taking}, APODA~\cite{yang2020adversarial}, PatchAlign~\cite{tsai2019domain}, DT~\cite{wang2020differential} and MRNet~\cite{zheng2019unsupervised}. Another line of works focuses on mining the target-domain knowledge via noisy labels and deploys the pseudo label learning, such as CBST~\cite{zou2018unsupervised}, MRKLD~\cite{zou2019confidence} and  Uncertainty~\cite{zheng2020unsupervised}, which generally achieves higher performance. 
As shown in Table~\ref{table:gtacity} and Table~\ref{table:syncity}, the proposed method yields consistent improvement in both kinds of works. The competitive performance has been achieved by combining AdaBoost Student with the Uncertainty~\cite{zheng2020unsupervised}. Specifically, we have achieved $50.9\%$ and $50.4\%$ mIoU on GTA5$\rightarrow$Cityscapes and SYNTHIA$\rightarrow$Cityscapes, respectively. Furthermore, the pre-class IoU results also suggest that the proposed method has achieved relatively good results on several classes of small-scale objectives, including Pole, Wall, and Fence.

\noindent\textbf{Cross-city.} We also evaluate the proposed method on one new cross-city benchmarks, \ie, Cityscapes $\rightarrow$ Oxford RobotCar (see Table~\ref{table:oxford}). Both Cityscapes and Oxford RobotCar are collected from the realistic scenarios but in different cities and weather. The proposed method also has achieved the competitive result of $75.2\%$ mIoU accuracy and showed a consistent performance improvement with other existing methods.

\noindent\textbf{Qualitative Results.} Furthermore, we provide  the segmentation results in Figure~\ref{fig:result}, 
%\textcolor{red}{5}  %(Due to the spatial limitation, we put the figure to the supplementary material)
which also verifies the effectiveness of the proposed method. Comparing to the segmentation predictions of vanilla MRNet~\cite{zheng2019unsupervised}, MRNet + Ours generally rectifies the object details, and achieves more robust outputs on the target domain. We apply the white dash boxes to highlight the different predictions. In particular, the proposed method is more robust to the minority categories, such as lights, poles, and traffic signs.

\section{Further Analysis and Discussions}~\label{sec:further}
\noindent\textbf{Network Structure.}
We also verify the proposed method on a relatively vanilla network, \ie, VGG16~\cite{simonyan2014very}. Following previous works, we do not introduce BN (Batch Normalization) layers in the backbone network. 
We re-implement MRNet~\cite{zheng2019unsupervised}, which has achieved 25.5\% mIoU on GTA5 $\rightarrow$ Cityscapes. 
As shown in Table~\ref{table:vgg}, we can observe two main points. First, the proposed method is complementary to the existing works, \eg, MRNet~\cite{zheng2019unsupervised}, and MRNet~\cite{zheng2019unsupervised} + Ours achieves +14.0\% mIoU improvements. Second, we also achieve one competitive result 39.5\% mIoU with other existing methods based on VGG16.

\setlength{\tabcolsep}{20pt}
\begin{table}[tb]
	\caption{
		Ablation Study on the effectiveness of the proposed method on a relatively vanilla network, \ie, VGG16~\cite{simonyan2014very} on GTA5 $\rightarrow$ Cityscapes. $^{*}$: We re-implement MRNet~\cite{zheng2019unsupervised} with VGG16.
	}
		\label{table:vgg}\vspace{-.1in}
\footnotesize
	\centering
	\begin{tabular}{l|c|c}
		\shline
		Method & Backbone & mIoU \\
		\hline
		Curriculum DA~\cite{zhang2017curriculum} & VGG16 & 28.9 \\
		Wang~\etal~\cite{wang2019weakly}& VGG19 & 33.1 \\
		CyCADA~\cite{hoffman2018cycada} & VGG16 & 35.4 \\
		ASA~\cite{zhou2020affinity} & VGG16 &35.6 \\ 
		CBST-SP~\cite{zou2018unsupervised} & VGG16 & 36.1 \\
		DCAN~\cite{wu2018dcan} & VGG16 & 36.2 \\
		SP-Adv~\cite{SHAN2020125} & VGG16 & 36.5 \\
		LSD~\cite{sankaranarayanan2018learning}
		& VGG16 & 37.1 \\
		\hline
		MRNet$^*$~\cite{zheng2019unsupervised} & VGG16 & 25.5 \\
		MRNet$^*$~\cite{zheng2019unsupervised} + Ours & VGG16 & 39.5 \\
		\shline
	\end{tabular}
\end{table}

\noindent\textbf{Aggregation with Momentum.} Due to the prediction fluctuation on the target domain, we can not foreknow which snapshot is reliable. We also run MRNet+Ours with momentum strategy~\cite{sutskever2013importance} on GTA5 $\rightarrow$ Cityscapes. Two momentum values $0.9$ and $0.5$ are explored to keep old weights. The aggregated model only achieves $46.4\%$ and $45.3\%$, respectively, which is inferior to the weight averaging. It verifies that the weight averaging trade-off is sub-optimal yet effective. In this paper, we leverage the weight averaging strategy, but it is still open to other aggregation alternatives.

\noindent\textbf{Focal Loss.} The focal loss is designed for supervised learning, which is modified from the cross-entropy loss~\cite{lin2017focal}. For domain adaptation, we do not have ground-truth target-domain labels. Therefore, we apply the focal loss on the source-domain data to train the model. We re-implement the focal loss, and apply the default gamma=2 as the original paper suggested. However, the result of Ours+Focal Loss (47.87\% mIoU) is not better than Ours. We think that the focal loss makes the model focus on the ``hard’’ samples in the source domain rather than target domain, which compromises the final performance. In the future, we will further study to fuse the focal loss in other formats on the target domain. 

\noindent\textbf{\emph{vs.} Mean Teacher (MT).} 
Similarly, the existing method, \ie, Mean Teacher (MT)~\cite{tarvainen2017mean} also preserves one weight-average model during the training process.  We apply the same strategy, including iteration numbers, to update the teacher model and re-implement  Mean Teacher. The main difference between the proposed method and Mean Teacher (MT) is twofold: First, we adopt the adaptive sampler, which enables interactions between learned models and the data sampler. In this way, the proposed method leads to more complementary snapshots during training for the student aggregation. Second, Mean Teacher needs the teacher prediction in every iteration. It demands extra model inference to calculate the kl-divergence loss (costs extra time and memory) and forces the student model and the weight-average teacher model to predict the same distribution given the input, which also limits learning complementary student models. In contrast, the proposed method does not introduce such an objective and performs more efficiently and effectively with the spirit of AdaBoost~\cite{freund1996experiments}. As a result, the proposed method outperforms MT (Student) 46.6\% and MT (Teacher) 47.5\% mIoU (see Table~\ref{table:meanteacher}). 

\setlength{\tabcolsep}{20pt}
\begin{table}[tb]
	\caption{
		Comparison with Mean Teacher (MT)~\cite{tarvainen2017mean}. 
	}\vspace{-.1in}
\label{table:meanteacher}
\footnotesize
	\centering
	\begin{tabular}{l|c}
		\shline
		Method & mIoU \\
		\hline
		MRNet~\cite{zheng2019unsupervised} & 45.5 \\
		\hline
		MRNet~\cite{zheng2019unsupervised} + MT (Student)& 46.6 \\
		MRNet~\cite{zheng2019unsupervised} + MT (Teacher)& 47.5  \\
		MRNet~\cite{zheng2019unsupervised} + Ours & 49.0 \\
		\shline
	\end{tabular}
\end{table}

\begin{table}[t]
%\scriptsize
%\small
\caption{Ablation study of Adaboost Student on GTA5+SYNTHIA$\rightarrow$Cityscapes. We adopt MRNet~\cite{zheng2019unsupervised} as baseline, and study the impact of the adaptive sampler and the student aggregation alone. The results also suggest that the adaptive sampler does not improve the single model performance but helps complementary model learning, facilitating the final ``weak'' student model aggregation.
} \label{table:multi}
\vspace{-.2in}
\footnotesize
\begin{center}
%\resizebox{\linewidth}{!}
{
\setlength{\tabcolsep}{3pt}
\begin{tabular}{l|c|c|c}
\shline
Method &  Adaptive & Student & mIoU \\
  & Sampler & Aggregation & \\
\hline
MRNet~\cite{zheng2019unsupervised} & & & 47.6 \\
$w$ Adaptive Sampler  & $\checkmark$ & & 48.4 \\
$w$ Student Aggregation  & & $\checkmark$ & 50.6 \\
MRNet~\cite{zheng2019unsupervised} + Ours  & $\checkmark$ & $\checkmark$ & 50.8 \\
\shline
\end{tabular}}
\end{center}
%\vspace{-.1in}
\end{table}

\noindent\textbf{Multi-Source Domains.} We add a new setting on  GTA5 + SYNTHIA$\rightarrow$Cityscapes with two source domains to evaluate the proposed method. With multi-source domain data, the model can be trained more robust to the unlabelled target environment as well as more diverse between single snapshots. As shown in Table~\ref{table:multi}, we could observe three points: (1) With more source-domain data, the model arrives at a better basic result of 47.6\% compared to 45.5\% (only GTA5). (2) The model with the adaptive sampler (48.4\%) is still competitive with the single snapshot on baseline (47.6\%).  (3) With better single snapshots, our method achieves the best performance of 50.8\% mIoU.

\setlength{\tabcolsep}{20pt}
\begin{table}[tb]
	\caption{
		Comparison with Mean Teacher (MT)~\cite{tarvainen2017mean} from 10 runs on CIFAR-10 using 4000 labels. $^*$: We re-run the offical code of Mean Teacher with a slight better performance.
	}\vspace{-.1in}
\label{table:cifar10}
\footnotesize
	\centering
	\begin{tabular}{l|c}
		\shline
		Method & top1 error ($\%$) \\
		\hline
		Supervised Learning~\cite{gastaldi2017shake} & 2.86 \\ 
		\hline
		VAT~\cite{miyato2018virtual} & 10.55 \\
		CT-GAN~\cite{wei2018improving} & 9.98 $\pm$ 0.21 \\
		Mean Teacher~\cite{tarvainen2017mean} & 6.28 $\pm$ 0.15 \\
		Mean Teacher~\cite{tarvainen2017mean}$^*$ & 6.14 $\pm$ 0.24 \\
		Ours & 6.05 $\pm$ 0.12 \\
		\shline
	\end{tabular}
\end{table}

\noindent\textbf{Semi-supervised Learning.} We follow MeanTeacher~\cite{tarvainen2017mean} and modify the official code on Cifar-10 as our baseline~\footnote{\url{https://github.com/layumi/Cifar10-Adaboost}}. The MeanTeacher backbone also has two classifiers, one for classification and the other for regressing the teacher prediction. Therefore, the proposed method can directly leverage the difference between two classifiers as the prediction variance to update the data sampler. As shown in Table~\ref{table:cifar10}, the proposed method further improves the semi-supervised learning result (10 runs with 4000 labeled data) from 6.28\% $\pm$ 0.15 top-1 error (reported in MeanTeacher) to 6.05\% $\pm$ 0.12 top-1 error. 

\section{Conclusion}~\label{sec:conclusion}
We identify one practical challenge in deciding the early-stopping time of the scene segmentation domain adaptation task. 
To address this limitation, we propose an efficient bootstrapping solution, AdaBoost Student, to learn the scalable model by aggregating ``weak'' models. 
The main idea underpinning AdaBoost Student is to leverage the model discrepancy during training. Specifically, AdaBoost Student adopts one adaptive data sampler and explicitly facilitates learning complementary models in a coherent manner. Thus, AdaBoost Student can efficiently converge to one robust final model and prevent over-fitting. 
As a result, we have achieved consistent improvements and competitive results on three benchmarks, including two synthetic-to-real benchmarks and one cross-city benchmark. In the future, we will continue to investigate the usage of Adaboost Student and apply it to other fields, such as medical images, to obtain the model with good generalizability.

% references section
\bibliographystyle{IEEEtran}
\bibliography{egbib}

\begin{IEEEbiography}[{\includegraphics[width=1in,height=1.25in,clip,keepaspectratio]{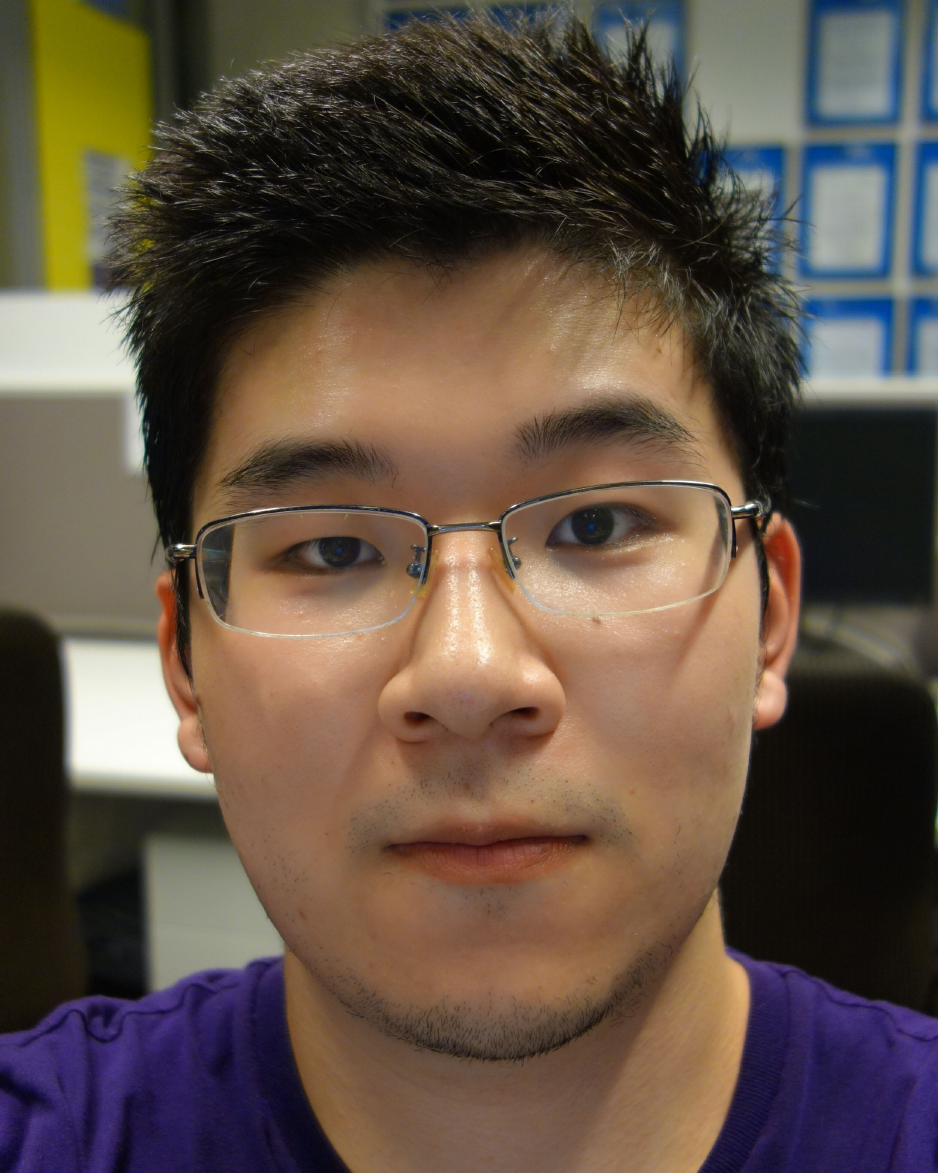}}]{Zhedong Zheng} received the Ph.D. degree from the University of Technology Sydney, Australia, in 2021 and the B.S. degree from Fudan University, China, in 2016. He is currently a postdoctoral researcher at NExT++, School of Computing, National University of Singapore. He was an intern at Nvidia Research (2018) and Baidu Research (2020). His research interests include robust learning for image retrieval, generative learning for data augmentation, and unsupervised domain adaptation.
\end{IEEEbiography}
\vfill
\begin{IEEEbiography}[{\includegraphics[width=1in,height=1.25in,clip,keepaspectratio]{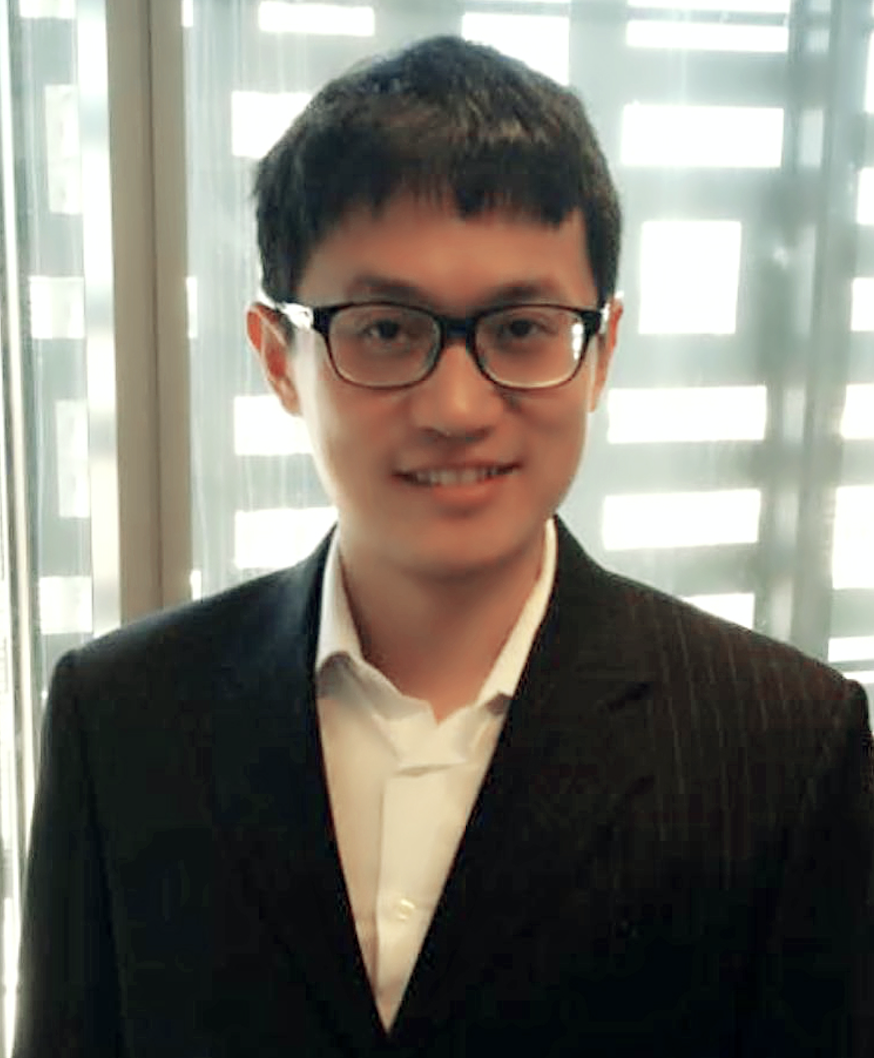}}]{Yi Yang} received the Ph.D. degree in computer
science from Zhejiang University, China, in 2010. He is currently a professor with the College of
Computer Science and Technology, Zhejiang University, China. He was a professor with University of Technology Sydney, Australia. Prior to joining UTS, he was a postdoctoral researcher with the School of Computer Science, Carnegie Mellon University, Pittsburgh, PA, USA. His current research interest includes machine learning and its applications to multimedia content analysis and computer vision, such as multimedia analysis and video semantics understanding.
\end{IEEEbiography}
\end{document}